\documentclass[10pt,twocolumn,letterpaper]{article}

\usepackage{cvpr}
\usepackage{times}
\usepackage{epsfig}
\usepackage{graphicx}
\usepackage{amsmath}
\usepackage{amssymb}
\usepackage{cite}
\usepackage{algorithm}
\usepackage{algorithmic}
\usepackage{array}
\usepackage{url}
\usepackage{stfloats}
\usepackage{color}
\usepackage{eqparbox}

\usepackage{multirow}
\usepackage{epstopdf}
\usepackage{booktabs}
\usepackage{bigstrut}
\usepackage{diagbox}
\usepackage{balance}
\usepackage{rotating}

\usepackage[pagebackref=true,breaklinks=true,letterpaper=true,colorlinks,bookmarks=false]{hyperref}

\cvprfinalcopy 

\ifcvprfinal\fi
\begin{document}

\title{Learning from Millions of 3D Scans for Large-scale 3D Face Recognition \\ (\textcolor{blue}{\textbf{This the preprint of the paper published in CVPR 2018}})} 

\author{Syed Zulqarnain Gilani {\hspace{4mm}} Ajmal Mian\\
	School of Computer Science and Software Engineering,\\
	The University of Western Australia\\
	{\tt\small \{zulqarnain.gilani,ajmal.mian\}@uwa.edu.au}
}
\date{}
\maketitle

\begin{abstract}
Deep networks trained on millions of facial images are believed to be closely approaching human-level performance in face recognition. However, open world face recognition still remains a challenge. Although, 3D face recognition has an inherent edge over its 2D counterpart, it has not benefited from the recent developments in deep learning due to the unavailability of large training as well as large test datasets. Recognition accuracies have already saturated on existing 3D face datasets due to their small gallery sizes. Unlike 2D photographs, 3D facial scans cannot be sourced from the web causing a bottleneck in the development of deep 3D face recognition networks and datasets. In this backdrop, we propose a method for generating a large corpus of labeled 3D face identities and their multiple instances for training and a protocol for merging the most challenging existing 3D datasets for testing. We also propose the first deep CNN model designed specifically for 3D face recognition and trained on 3.1 Million 3D facial scans of 100K identities. Our test dataset comprises 1,853 identities with a single 3D scan in the gallery and another 31K scans as probes, which is several orders of magnitude larger than existing ones. Without fine tuning on this dataset, our network already outperforms state of the art face recognition by over 10\%. We fine tune our network on the gallery set to perform end-to-end large scale 3D face recognition which further improves accuracy. Finally, we show the efficacy of our method for the open world face recognition problem.
\end{abstract}

\vspace{-0mm}
\section{Introduction}
\label{sect_intro}
\vspace{-0mm}
Face recognition, being a highly non-intrusive biometric~\cite{bowyer2006}, is fast becoming the tool of choice~\cite{mian2012} in the domains of  surveillance (for example, border control, suspect tracking, identification), security (for example, system login, banking, file encryption) and entertainment (for example, human computer interaction, 3D animation, virtual reality). Advancements in Deep Learning have brought about revolutionary improvements in various computer vision tasks where CNN based face recognition  is claimed to have surpassed human performance~\cite{taigman2014}. However, the recent MegaFace challenges~\cite{nech2017,kemelmacher2016} have shattered this myth, revealing that face recognition is still an unsolved problem. 

\begin{table}[t]
	\centering
	\scriptsize
	\setlength{\tabcolsep}{1.25pt}
	\renewcommand{\arraystretch}{1.25}
	\caption{State-of-the-art 2D face recognition networks are trained on millions of images and tested on thousands of identities. However, 3D face recognition algorithms are tested on just a few hundred identities. The proposed FR3DNet is trained on 3.1M 3D scans and tested on 1.85K identities.}
	\vspace{2mm}
	\begin{tabular}{|c|l|c|cc|ccc|c|}
		\hline
		{\textbf{Modal-}} & \multicolumn{1}{c|}{{\textbf{Model} \textbackslash}} & \textbf{Input} & \multicolumn{2}{c|}{\textbf{Training}} & \multicolumn{3}{c|}{\textbf{Testing}} & {\textbf{NW }} \\
		
		\cline{4-8}     \textbf{ity}       &   \multicolumn{1}{c|}{\textbf{Technique}} & \textbf{Size}  & {\textbf{IDs}} & {\textbf{Scans}} & \multicolumn{1}{c}{{\textbf{IDs}}} & \multicolumn{1}{c}{{\textbf{Scans}}} & {\textbf{Dataset}} & \textbf{Param} \\
		
		\hline
		\multirow{4}[2]{*}{2D } & VGG-Face~\cite{parkhi2015} & 224 $\times$ 224 &2.6K & 2.6M  & 5K    & 13K   & LFW   & 134M \\
		& DeepFace~\cite{taigman2014} & 152 $\times$ 152 & 4K & 4.4M  & 5K    & 13K   & LFW   & 120M \\
		& FaceNet~\cite{schroff2015} & 220 $\times$ 220 & 8M    & 200M  & 5K    & 13K   & LFW   & 140M \\
		& MF2~\cite{nech2017} & - &672K  & 4.7M  & 690K  & 1M    & MegaFace & - \\ \hline
		
		\multirow{3}[4]{*}{3D} & MMH~\cite{mian2007} & - & -     & -     & \textcolor{red}{0.46K} & \textcolor{red}{4K}    & FRGCv2 & - \\
		& K3DM~\cite{gilani2017b} & - &-     & -     & \textcolor{red}{0.46K} & \textcolor{red}{4K}      & FRGCv2 & - \\
		& Kim \etal\cite{kim2017} & 224 $\times$ 224 & \textcolor{red}{0.7K}   & \textcolor{red}{123K}   & \textcolor{red}{0.1K}    & \textcolor{red}{4.6K}   & Bosphorus & 140M \\ \hline \hline
		\textbf{ 3D}    & \textbf{FR3DNet} & 160 $\times$ 160 & \textcolor{blue}{100K}  & \textcolor{blue}{3.1M}    & \textcolor{blue}{1.85K} & \textcolor{blue}{31K} & {LS3DFace} & \textbf{29M} \\ \hline
	\end{tabular}%
	\label{tab:introtab}%
	\vspace{-0mm}
\end{table}%

Two-dimensional face recognition using CNNs on conventional photographs has shown remarkable performance on benchmarks like LFW~\cite{huang2007} and Janus~\cite{klare2015}. One of the main factors for this accomplishment is the ability of CNNs to learn from massive training data which is readily available. For instance, FaceNet~\cite{schroff2015} was trained on 200M textured images of 8M identities while VGG-Face~\cite{parkhi2015} used 2.6M photos of 2,622 distinct subjects for training. Despite this phenomenal performance and availability of data, 2D face recognition is challenged by changes in illumination, pose  and scale~\cite{abate2007}. Furthermore, facial texture is not always stable for identities as it can change with make up. On the other hand, 3D face recognition has the potential to address these shortcomings. Although this modality in face recognition is gaining popularity~\cite{gilani2017a,gilani2017b,mian2008,al2009,blanz2003,berretti2013,li2014}, literature survey shows that there is no deep CNN designed specifically for 3D face recognition. This is primarily because of the lack of huge amounts of 3D training and test data. 3D face data cannot be obtained by crawling the web~\cite{nech2017,kemelmacher2016,parkhi2015} and it requires great efforts to collect a respectable sized dataset. For instance, the largest publicly available 3D face dataset, ND-2006~\cite{faltemier2007} (a superset of FRGCv2~\cite{phillips2005}) has only 13,540 scans of $888$ unique identities and took over two years to collect.

The problem of addressing the dearth of labeled 3D face data for training CNNs has been addressed through data augmentation. This is either done by creating synthetic faces from an existing 3D face model~\cite{dou2017,richardson2016} or by manipulating the facial appearance of existing data by introducing expressions~\cite{kim2017,masi2016}. The former method is restricted to the  linear space of the specific model resulting in faces with confined shape variations. The latter method only generates more scans per subject without increasing the number of unique identities in the data. In this paper, we present a  technique for data augmentation that introduces non-linear heterogeneous variations in 3D shape, facial expressions, pose and occlusions to generate a training dataset of 3.1M 3D scans of 100K unique identities. The closest numbers in literature~\cite{kim2017} for fine tuning VGG-Face on depth images are 127K scans of 700 identities, several orders of magnitude lower than ours (See Table~\ref{tab:introtab} for details).

Another notable challenge to face recognition systems is the need for large-scale of test data. Recognition accuracies on small datasets like LFW ($99.6\%$~\cite{schroff2015}) and FRGCv2 ($98.7\%$~\cite{gilani2017b}) have already saturated indicating the need for larger gallery sizes as it is well known that increasing the gallery size degrades the face recognition performance. The MegaFace Challenges~\cite{nech2017,kemelmacher2016} show that the performance of even the best 2D face recognition networks drop significantly when the gallery size increases. The identification accuracy of VGG network with triplet loss reduced by more than $20\%$ on FaceScrub when only $10^2$ distractors were added to the gallery set~\cite{nech2017}. FaceNet~\cite{schroff2015} behaved similarly when one million distractors were added to the gallery~\cite{kemelmacher2016}. Literature has no such statistics for 3D face recognition as large-scale 3D face recognition has never been attempted. Absence of large 3D face datasets with huge galleries is the prime reason for this massive gap in research. While millions of 2D face datasets have been generated by crawling the Internet~\cite{nech2017,kemelmacher2016,guo2016}, 3D domain still depends on physical collection of data from real subjects. 

We present a unique solution by merging the most challenging publicly available 3D face datasets for large-scale face recognition testing. Our gallery consists of 1,853 identities while the probe set contains 31,860 3D scans of these individuals. Through extensive experiments, we show how existing methods and CNN models perform on this large scale dataset. We use the challenging protocol of a single sample per identity in the gallery as, most often than not, this would be the case in practical real world scenarios. Note that in the domain of 3D face recognition,  the largest dataset (FRGCv2~\cite{phillips2005}) on which results have mostly been reported has only $466$ identities in the gallery.

Apart from data, the recognition algorithm itself is a very important component. The literature contains a variety of state-of-the-art deep CNN architectures  for 2D face recognition~\cite{sun2014,schroff2015,parkhi2015,resnethe2016}. Using networks trained on 2D images to perform 3D face recognition is simplistic and suboptimal as 3D data has its own peculiarities defined by the underlying shape and geometry. To the best of our knowledge, there is no deep network designed specifically for 3D face recognition. We cover this research gap and propose a Deep 3D Face recognition Network coined \textit{FR3DNet} (pronounced frednet) suited for 3D face data and trained from scratch on 3.1M 3D faces. We also analyze the affects of input image sizes and suitability of kernel sizes for 3D faces.

In a nutshell, our contributions are as follows: (1) \textit{Training Data}: We present a method for generating a large corpus of labeled 3D face data for training CNNs. Our dataset contains 3.1M 3D scans of 100K identities highly rich in shape variations. Our training data does not include the public datasets. (2) \textit{Large-scale Test Data}: Owing to the limitations of physically collecting huge 3D datasets, we merge the most challenging existing public 3D face datasets and propose a protocol for large-scale face recognition using a single sample per identity in the gallery. The test data contains 31,860 3D scans of 1,853 identities. To the best of our knowledge, this is the largest gallery size of 3D faces on which ~ face ~ recognition ~ results ~ have ~ ever  been  reported. (3) \textit{Deep 3D Face Recognition Network (FR3DNet)}: We propose the first ever deep CNN designed specifically for 3D face recognition and trained on 3.1M 3D faces. We fine tune \textit{FR3DNet} on the 1,853 gallery identities in our large-scale dataset and achieve an end-to-end Rank-1 recognition rate of $98.74\%$ on 27K probes, significantly outperforming the state-of-the-art on constituent datasets. The trained and end-to-end fine tuned \textit{FR3DNet} will be made public.

\vspace{0mm}
\section{Related Work}
\vspace{-1mm}

Face recognition is one of the most researched topics in Computer Vision and many detailed surveys exist~\cite{zhou2014, soltanpour2017,patil2015,bowyer2006}. Here, we present the most relevant works to this paper and divide them into conventional methods which use hand crafted local and global features, deep learning based methods which are mainly based on various CNN architectures and data augmentation methods which focus on the problem of limited training data for learning.

\vspace{2mm}
\noindent{\textit{\textbf{Conventional Methods for 3D Face Recognition:} }} These methods can be grouped into local or global descriptor based techniques~\cite{bowyer2006,abate2007} where the latter also include 3D morphable model based methods. Local descriptor based techniques match local 3D point signatures derived from the curvatures, shape index and/or normals. For instance, Mian \etal~\cite{mian2008} proposed a highly repeatable keypoint detection algorithm for 3D facial scans. They fused the 3D keypoints with 2D Scale Invariant Feature Transform (SIFT) to develop multimodal face recognition. However, the keypoint detection method and features were both sensitive to facial expressions. For robustness to facial expressions, Mian \etal~\cite{mian2007} proposed a parts based multimodal hybrid method (MMH) which exploited local and global features in the 2D and 3D modalities. A key component of their method was a variant of the ICP \cite{besl1992} algorithm which is computationally expensive due to its iterative nature. Gupta \etal~\cite{texasgupta2010} matched the 3D Euclidean and geodesic distances between pairs of fiducial landmarks to perform 3D face recognition. Berretti \etal~\cite{berretti2013} represented a 3D face with multiple meshDOG keypoints and local geometric histogram descriptors while Drira \etal~\cite{drira2013} represented the facial surface by radial curves emanating from the nosetip.

Model based methods construct a 3D morphable face model and fit it to each probe face. Face recognition is performed by matching the model parameters to those in the gallery. Gilani \etal~\cite{gilani2017b} proposed a keypoint based dense correspondence model and performed 3D face recognition by matching the parameters of a statistical morphable model called K3DM. Blanz \etal~\cite{blanz2003,blanz2007} used the parameters of their 3DMM~\cite{blanz1999} for face recognition. Passalis \etal~\cite{passalis2005} proposed an Annotated Face Model (AFM) based on an average facial 3D mesh. Later, Kakadiaris \etal~\cite{kakadiaris2007} proposed elastic registration using this AFM and performed 3D face recognition by comparing the wavelet coefficients of the deformed images obtained from morphing. Model fitting algorithms can be computationally expensive and do not perform well on large galleries as shown in our results.  

Both local and global techniques were tested on individual 3D datasets, the largest one being FRGCv2 with a gallery size of $466$ identities. To the best of our knowledge, none of the conventional methods have performed large-scale 3D face recognition.

\vspace{2mm}
\noindent{\textit{\textbf{Deep Learning:} }}Akin to progress in other applications of computer vision, deep learning has given a quantum jump in 2D face recognition. Three years ago, Facebook AI group proposed a nine-layer DeepFace model~\cite{taigman2014} mainly consisting of two convolutional, three locally-connected and two fully-connected (FC) layers. The network was trained on 4.4M 2D facial images of 4,030 identities and achieved an accuracy of $97.35\%$ on the benchmark LFW~\cite{huang2007} dataset which is $27\%$ higher than the previous state of the art. This was followed by Google Inc., a year later, with FaceNet~\cite{schroff2015} based on eleven convolutional and three FC layers. The distinction of this network was its training dataset of 200M face images of 8M identities and a triplet loss function. The authors reported face recognition accuracy of $98.87\%$ on LFW. DeepFace and FaceNet were both trained on private datasets which are not available to the broader research community. Consequently, Parkhi \etal~\cite{parkhi2015} proposed a method for crawling the web to collect a face database of 2.6M 2D images from 2,622 identities and presented the VGG-Face model comprising of $16$ convolutional and three FC layers. Despite training on a smaller dataset, the authors reported face recognition accuracy of $98.95\%$ on the LFW dataset. However, recently the MegaFace Challenges~\cite{nech2017,kemelmacher2016} claimed that the existing 2D benchmark datasets have reached saturation and proposed adding millions of faces to the galleries of these datasets to match the real world scenarios. They showed that the face recognition accuracy of state-of-the-art 2D networks dropped by more that $20\%$ when just a few thousand distractors were added to the gallery of public face recognition benchmark datasets. The take away for the 3D domain is that CNNs on 2D data perform best when they learn from massive training sets and are particularly designed for the 2D modality, and yet, their real performance can be validated only when they are tested with large gallery sizes.

To the best of our knowledge, only Kim \etal~\cite{kim2017} have presented deep 3D face recognition results. They reported results on three public datasets after fine tuning the VGG-Face network~\cite{parkhi2015} on 3D depth images.  They used an augmented dataset of 123,325 depth images to fine-tune the VGG-Face network and then tested it on the Bosphorus~\cite{savran2008}, BU3DFE~\cite{yin2006} and 3D-TEC (twins)~\cite{twinsvijayan2011} datasets individually. Except for the Bosphorus dataset, their results do not outperform the state-of-the-art conventional methods. Moreover, they have not reported results on the challenging FRGCv2 dataset and their fine-tuned model is not publicly available.

\noindent{\textit{\textbf{Data Augmentation:} }} 
Dou \etal~\cite{dou2017} and Richardson \etal~\cite{richardson2016} generated thousands of synthetic 3D images for face reconstruction using BFM~\cite{paysan2009}, AFM~\cite{kakadiaris2007} and 3DMM~\cite{blanz1999}. This method generates 3D faces within the linear space of a specific statistical face model. The faces generally have a variation of $\pm3$ standard deviations from the model mean with highly smooth surfaces. Gilani \etal~\cite{gilani2017a} generated synthetic images using a similar approach. However, these images were used to train a 3D landmark identification network. Kim \etal~\cite{kim2017} fitted the BFM~\cite{paysan2009} to $577$ identities of FRGCv2~\cite{phillips2005} database and induced $25$ expressions in each identity. They also introduced minor pose variations between $\pm10^\circ$ in yaw, pitch and roll for each original scan. To simulate occlusions, the authors introduced eight random occlusion patches to each 2D depth map to increase the dataset to 123,325 scans. This method only increases the intra-person variations without augmenting the number of identities, which in this case remained $577$. 

\begin{figure}[tb]
	\centering
	\includegraphics[trim = 0pt 0pt 0pt 0pt, clip, width=1\linewidth]{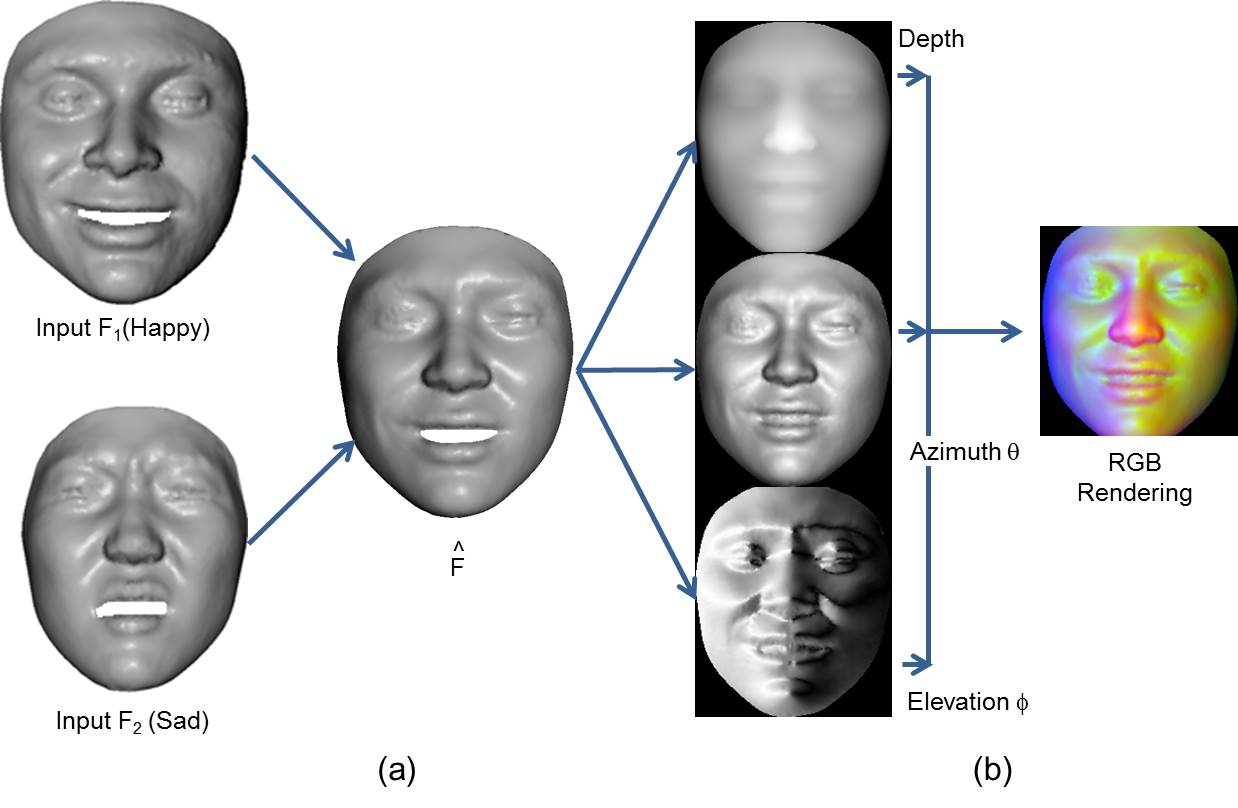}
	\caption{\small{(a) Our data generation process. Notice the non-linearity introduced in the new face while at the same time preserving the high frequency shape variations. (b) Data preparation for input to our \textit{FR3DNet}.  }}
	\label{fig:datagen1}
	\vspace{-0mm}
\end{figure}

\begin{figure}[tb]
	\centering
	\includegraphics[trim = 0pt 0pt 0pt 0pt, clip, width=1\linewidth]{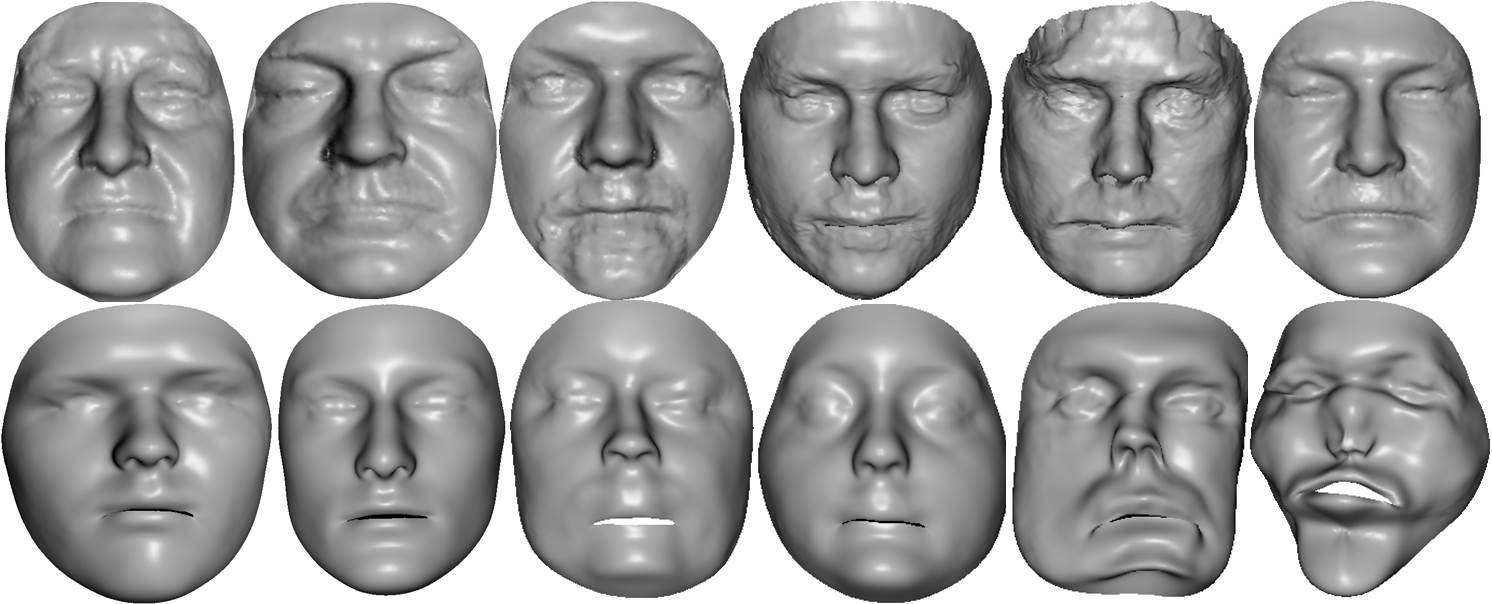}
	\caption{\small{Example 3D faces generated by our method (row 1) and a statistical model~\cite{paysan2009} (row 2). The same identities were used for generating faces for both techniques. The 3D faces from our method look more realistic and have richer shape variations, especially around high curvature regions.}}
	\label{fig:output_comp}
	\vspace{-6mm}
\end{figure}

\vspace{0mm}
\section{Proposed Data Generation for Training}
\vspace{-1mm}
We use 3D facial scans of 1,785 individuals (a propriety dataset) who were participants of various studies in our institution to train our deep network. The number of identities in this dataset is larger than any 3D dataset but still not sufficient for deep learning. Inspired by the recent works of Gilani~\etal~\cite{gilani2017b}, we establish dense correspondence over 15K 3D vertices on the faces from this dataset, using the keypoints based algorithm. The goal now is to grow the dataset by generating faces from the space spanned by pairs of densely corresponding real 3D faces of distinct identities. To ensure that the identities in the pair are as ``distinct'' as possible, we select the face pair with the maximum non-rigid shape difference. Let the faces be represented by $\mathbf{F}_i=[x_p,y_p,z_p]^T$, where $i=1,\ldots,N$, $p=1,\ldots,P$; $N=1,000$ and $P=$15,000. 
The shape difference between faces $\mathbf{F}_i$ and $\mathbf{F}_j$ is defined as
\vspace{0mm}
\begin{equation}
\vspace{-1mm}
\mathbf{D}(i,j) = \dfrac{\gamma_{ij}+\gamma_{ji}}{2} ,
\label{bendingenergy}
\end{equation}
where, $\gamma_{ij}$ is the amount of bending energy required to deform 3D face $\mathbf{F}_i$ to face $\mathbf{F}_j$. Extending the 2D thin-plate spline model~\cite{bookstein1989} to our case, we calculate the bending energy as, 
$\gamma(i,j) = \mathbf{x}^T\boldsymbol{\mathcal{B}}\mathbf{x}+\mathbf{y}^T\boldsymbol{\mathcal{B}}\mathbf{y}+\mathbf{z}^T\boldsymbol{\mathcal{B}}\mathbf{z}$
where $\mathbf{x}$,~$\mathbf{y}$ and $\mathbf{z}$ are the vectors containing the $x$, $y$ and $z$ coordinates of $P$ points in face $\mathbf{F}_j$ and $\boldsymbol{\mathcal{B}}$ is the bending matrix, which is defined as the  $P\times P$ upper left matrix of 
$\left[\begin{matrix}
\mathbf{K}~~\mathbf{S}
\\ \mathbf{S}^T~\mathbf{0}
\end{matrix}\right]^{-1}$. Here $\mathbf{K}(a,b)=||\mathbf{F}^a_i-\mathbf{F}^b_i||^2~ log||\mathbf{F}^a_i-\mathbf{F}^b_i||$ with $a,b=1,\ldots,P$, $\mathbf{S}=[\mathbf{1}, {\mathbf{x}^j}, {\mathbf{y}^j}, {\mathbf{z}^j}]$, and $\mathbf{0}$ is a $P\times4$ matrix of zeros.

We select 90,100 pairs of 3D faces with maximum shape difference $\mathbf{D}(i,j)$ from the possible $\binom N2 = 499,500$ pairs. Since the 3D faces in each pair are in dense correspondence to each other, a new face $\hat{F}$ is generated from the linear space of each pair $(i,j)$ as $\hat{F}= \frac{[x^p_i,y^p_i,z^p_i]^T+[x^p_j,y^p_j,z^p_j]^T}{2} $. The process is depicted in Figure~\ref{fig:datagen1}. 

It is important to note here that our proposed method is significantly different from generating synthetic faces from a statistical face model. Varying the parameters of a statistical model generates faces that are over smooth and devoid of details and high frequency shape variations because of the low dimensional space that is used to generate them. On the contrary, our synthetic faces are generated from high dimensional raw 3D faces. Furthermore, not all faces generated by statistical models are \textit{faces} unless strict constraints are imposed on the variation of the model parameters~\cite{mian2013}. Such constraints will further limit the variations in identities that can be generated from the model. Finally, faces generated from statistical models span the linear space of the model whereas our method introduces non-linearity in the generated identities by varying the expressions of the face pair used to generate $\hat{F}$. By interpolating between identities and expressions, we generate new identities that do not necessarily lie in the linear space of the original identities. This is illustrated in Figure \ref{fig:datagen1}. Thus, we can choose the most dissimilar faces generating new identities that have maximum inter-person variations. 
The differences in the two methods of face generation can be seen clearly in Figure~\ref{fig:output_comp}. Note that it is guaranteed that our method will never create  deformed un-realistic faces  like the ones generated by the statistical model (for example last two faces of bottom row).

The second source of 3D faces for our training data is a commercial software~\footnote{Singular Inversions, “Facegen Modeller”, www.facegen.com} that generates densely corresponded faces of varying facial shapes, ethnicities and expressions. We generate $300$ identities, each in four different expressions with three intensity levels and follow the protocol above to create 9,950 new identities from the 44,850 possible pairs. However, in this case we select the pairs of faces that are ``similar'' and have smaller inter-person distance as per definition in Equation~\ref{bendingenergy}. The motivation for placing this condition comes from real world scenarios where face recognition systems are required to recognize people who look quite identical, for example in extreme cases, identical twins or triplets. A face recognition system trained on identities that look similar would have the power to distinguish between probes that are very similar in shape. Note that there is still ample inter-person variation in the original pairs for our \textit{FR3DNet} to learn high level face identity features.

\begin{figure*}[t]
	\centering
	\includegraphics[trim = 0pt 0pt 0pt 0pt, clip, width=1\linewidth]{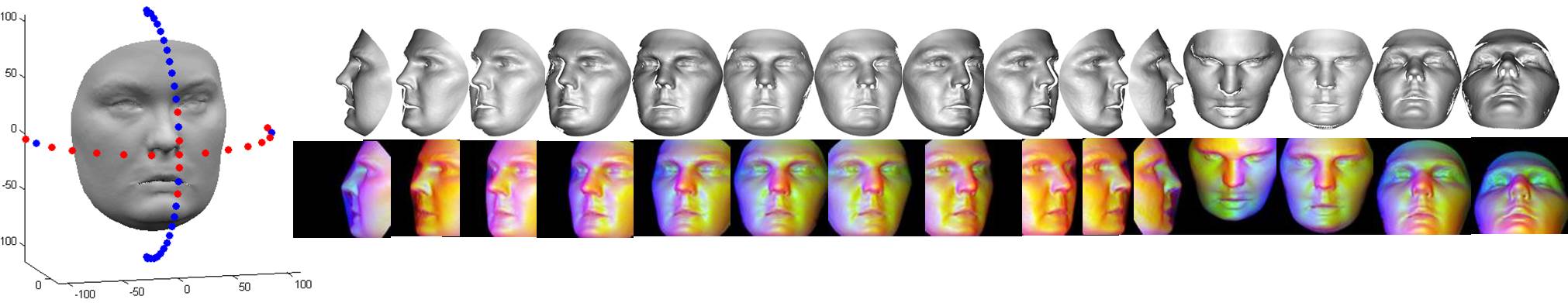}
	\caption{\small{Position of cameras on a hemisphere surrounding the 3D face and the $15$ poses generated as a result.}}
	\label{fig:posevar}
	\vspace{0mm}
\end{figure*}
Finally, we simulate pose variations and large occlusions in each 3D scan by deploying $15$ synthetic cameras on a hemisphere in front of the 3D face. The cameras are deployed in the range of $[-90^\circ,90^\circ]$ on the longitude and $[-30^\circ,30^\circ]$ on the latitude of the hemisphere; all at $15^\circ$ intervals. We do not deploy cameras at $-75^\circ$ and $75^\circ$ on the longitude. The self-occluded 3D points from the camera view point  are removed by applying the hidden point removal algorithm~\cite{katz2007}. Note that this step creates  missing data in varying amounts on each scan, thereby simulating realistic self occlusion. Figure~\ref{fig:posevar} depicts the placement of cameras and displays the output images. 

Our final training dataset consists of 3,169,275 facial scans from 100,005 identities (approx $31$ scans per identity). Table~\ref{tab:trgdata} gives details of the augmented 3D face dataset.

\section{\textit{FR3DNet}: Deep Network for 3D Face Recognition }
\label{sect_d3dfn}
\subsection{Training Data}
The 3D pointcloud of each scan in the training data is used to generate a three channel image. The first channel is the depth image which is generated by fitting a surface of the form $z(x,y)$ to the 3D pointcloud using the \textit{gridfit} algorithm~\cite{john2008}. The surface normals of the original pointcloud are calculated in spherical coordinates ($\theta, \phi$) where $\theta$, $\phi$ are the azimuth and elevation angles of the normal vector. Using a similar $x,y$ grid to the depth image, surfaces of the form $\theta(x,y)$ and $\phi(x,y)$ are fitted to the azimuth and elevation angles to make the second and third channels of the 3D image representation we used to train our network. The three channels are normalized on the 0-255 range and can be rendered as an RGB image. This image is  passed through a landmark identification network~\cite{gilani2017a} to detect the nosetip. With the face centered at the nosetip, we crop a square of $224\times 224$ pixels. This process is depicted in Figure~\ref{fig:datagen1}. The $224\times 224$ size is chosen for comparison with existing networks. These images are down-sampled to $160\times 160$ for use in our network.

\begin{figure}[t]
	\centering
	\includegraphics[trim = 0pt 0pt 0pt 0pt, clip, width=1\linewidth]{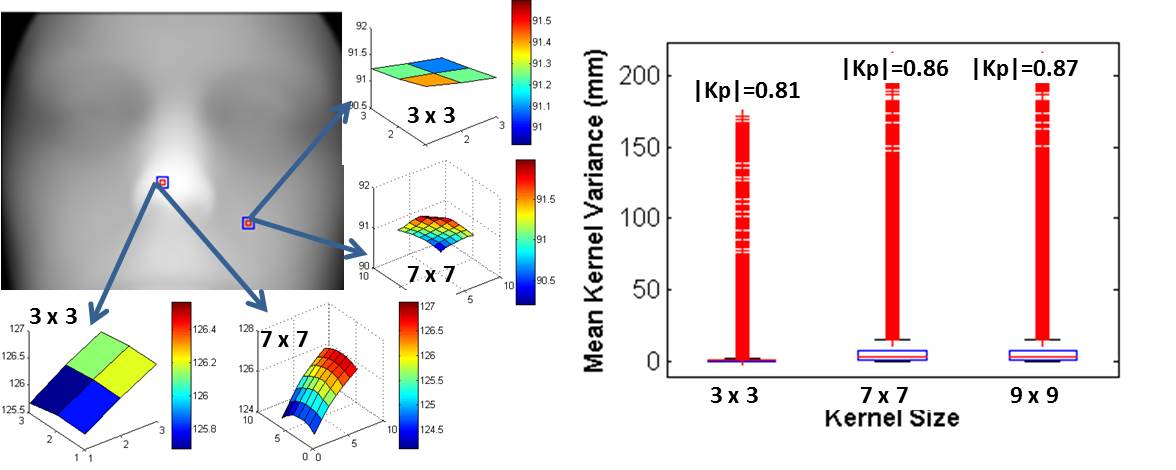}
	\caption{\small{Variation in 3D depth  frequency with different kernel sizes. Notice that patches of $3\times3$ are almost quasi-planar. $|K_p|$ denotes average number of keypoints per kernel size.}}
	\label{fig:kernelsize}
	\vspace{0mm}
\end{figure}

\begin{table}[htbp]
	\centering
	\scriptsize
	\setlength{\tabcolsep}{2.5pt}
	\renewcommand{\arraystretch}{1.25}
	\caption{Details of the dataset generated for training \textit{FR3DNet}.}
	\begin{tabular}{|l|r|c|c|r|}
		\hline
		\multicolumn{1}{|l|}{\textbf{Type}} & \textbf{IDs} & \textbf{Expressions} & \textbf{Poses} & \textbf{Total Scans} \\
		\hline
		{Dense Correspondence Model} & 90,100 & 2     & 15    & *1,680,900 \\
		{Real 3D Faces} & 1,785  & 1     & 15    & 26,775 \\
		{Synthetic} & 8,120  & 12    & 15    & 1,461,600 \\
		\hline
		\textbf{Total} & \textbf{100,005} & \textbf{12} & \textbf{15} & \textbf{3,169,275} \\
		\hline
		\multicolumn{5}{l}{*Randomly selected from 2,703,000 scans.}
	\end{tabular}%
	\label{tab:trgdata}%
	\vspace{0mm}
\end{table}%

\vspace{0mm}
\subsection{Network Architecture and Feature Extraction}
\vspace{0mm}
Inspired by the success of recent deep networks~\cite{simonyan2014,parkhi2015} in 2D face recognition,  we propose a deep convolutional neural network that is suited to 3D data. The VGG network was designed for 2D images which exhibit significant texture variations over small regions. In contrast, 3D facial surfaces are generally smooth and hence filters with larger kernel sizes would better suite this type of data. For example, Figure~\ref{fig:kernelsize} shows that surface patches of $7 \times 7$ contain more variation than patches of $3 \times 3$ and this is true even for the high curvature areas. This claim is empirically verified by calculating the average variance and average number of keypoints~\cite{mian2010} over kernel sizes of $3$, $7$ and $9$ in 10,000 3D images randomly selected from our training data. Average keypoints are calculated as the number of points on the 3D facial image that qualify as keypoints for a given kernel size, using the criterion in \cite{mian2010}, divided by the number of possible kernels of that size on the image. The average kernel variance and average number of keypoints per kernel size of $7 \times 7$ is significantly higher than size $3\times3$. Results depicted in Figure~\ref{fig:kernelsize} are compelling in favor of a kernel size of $7$ for our initial convolutional layers. 

\begin{figure}[tb]
	\centering
	\includegraphics[trim = 0pt 0pt 0pt 0pt, clip, width=1\linewidth]{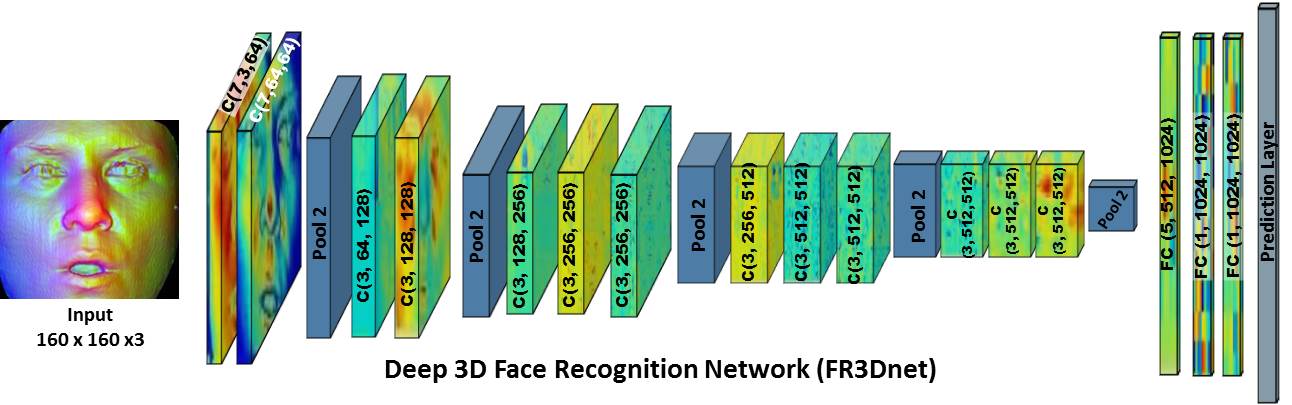}
	\caption{\small{Architecture of our proposed \textit{FR3DNet}. Every convolutional layer is followed by a rectifier layer.  }}
	\label{fig:fr3dnet}
	\vspace{-0mm}
\end{figure}

The skeleton architecture of our \textit{FR3DNet} follows~\cite{parkhi2015} but with a change in the \textit{conv} layers, details of which are given in Figure~\ref{fig:fr3dnet}. We aim to minimize the average prediction log-loss after the softmax layer by learning the parameters of a network designed to classify $N=$100,005 identities. After the network is trained, we remove the drop out layers. The embedded feature vectors of length 1,024 from FC7 can be used for face recognition  by minimizing the cosine distance between a probe and the gallery faces in the feature space. We also fine-tune the \textit{FR3DNet} on the gallery scans of the large-scale test data and denote it as  \textit{FR3DNet}$_{FT}$. 

\subsection{Implementation Details}
The input to our network is the $160 \times 160 \times 3$ image where the three channels correspond to depth, azimuth and elevation angles of the normal vector. We train the proposed \textit{FR3DNet} in MatConvNet~\cite{vedaldi15} with randomly selected $90\%$ scans of each identity in training and use the remaining scans for validation. We optimize the learning by Stochastic Gradient Descent (SGD) with standard L2 norm over the learned weights using mini batches of $20$ images. The model is regularized using dropout layers after FC6 and FC7 with $0.5$ rate. The learning rate was initially set to $0.01$ and reduced by a factor of $10$ after every $10$ epochs. The filter weights of each layer were initialized with parameters drawn from a Gaussian distribution with zero mean and a standard deviation adjusted using the Xavier's method~\cite{glorot2010}.  The network was trained for $50$ epochs. 
For fine-tuning \textit{FR3DNet}$_{FT}$, the network weights were frozen except for the last layer which was learnt with a rate of $0.01$.  Since the gallery contains only one 3D scan per identity, we render it from multiple viewpoints to generate more training data.

\begin{table}[bt]
	\centering
	\scriptsize
	\setlength{\tabcolsep}{3.0pt}
	\renewcommand{\arraystretch}{1.2}
	\caption{Details of the constituent datasets of \textit{LS3DFace}.}
	\vspace{2mm}
	\begin{tabular}{lllllll}
		\hline
		\multicolumn{1}{|l|}{\textbf{Name}} & \multicolumn{1}{c|}{\textbf{IDs}} & \multicolumn{1}{c|}{\textbf{Scans}} & \multicolumn{1}{l|}{\textbf{Expressions}} & \multicolumn{1}{l|}{\textbf{Pose}} & \multicolumn{1}{c|}{\textbf{Occlusion}} & \multicolumn{1}{l|}{\textbf{Scanner}} \\
		\hline
		\multicolumn{1}{|l|}{FRGCv2~\cite{phillips2005}} & \multicolumn{1}{c|}{466} & \multicolumn{1}{c|}{4,007} & \multicolumn{1}{c|}{Multiple} & \multicolumn{1}{c|}{$\pm15^\circ$} & \multicolumn{1}{l|}{None} & \multicolumn{1}{l|}{Laser} \\
		\multicolumn{1}{|l|}{BU3DFE~\cite{yin2006}} & \multicolumn{1}{c|}{100} & \multicolumn{1}{c|}{2,500} & \multicolumn{1}{c|}{6 $\times$4} & \multicolumn{1}{c|}{Frontal} & \multicolumn{1}{l|}{None} & \multicolumn{1}{l|}{Sterio} \\
		\multicolumn{1}{|l|}{Bosphorus~\cite{savran2008}} & \multicolumn{1}{c|}{105} & \multicolumn{1}{c|}{4,666} & \multicolumn{1}{c|}{7} & \multicolumn{1}{c|}{$\pm90^\circ$} & \multicolumn{1}{l|}{4 types} & \multicolumn{1}{l|}{Sterio} \\
		\multicolumn{1}{|l|}{GavabDB~\cite{gavabmoreno2004}} & \multicolumn{1}{c|}{61} & \multicolumn{1}{c|}{488} & \multicolumn{1}{c|}{Miltiple} & \multicolumn{1}{c|}{$\pm30^\circ$} & \multicolumn{1}{l|}{None} & \multicolumn{1}{l|}{Laser} \\
		\multicolumn{1}{|l|}{Texes FRD~\cite{texasgupta2010}} & \multicolumn{1}{c|}{118} & \multicolumn{1}{c|}{1,151} & \multicolumn{1}{c|}{Miltiple} & \multicolumn{1}{c|}{Frontal} & \multicolumn{1}{l|}{None} & \multicolumn{1}{l|}{Sterio} \\
		\multicolumn{1}{|l|}{BU4DFE~\cite{bu4yin2008}} & \multicolumn{1}{c|}{101} & \multicolumn{1}{c|}{3,030} & \multicolumn{1}{c|}{6 $\times$5} & \multicolumn{1}{c|}{Frontal} & \multicolumn{1}{l|}{None} & \multicolumn{1}{l|}{Sterio} \\
		\multicolumn{1}{|l|}{CASIA~\cite{casia}} & \multicolumn{1}{c|}{123} & \multicolumn{1}{c|}{4674} & \multicolumn{1}{c|}{6} & \multicolumn{1}{c|}{$\pm90^\circ$} & \multicolumn{1}{l|}{None} & \multicolumn{1}{l|}{Laser} \\
		\multicolumn{1}{|l|}{UMB DB~\cite{umbcolombo2011}} & \multicolumn{1}{c|}{143} & \multicolumn{1}{c|}{1,473} & \multicolumn{1}{c|}{4} & \multicolumn{1}{c|}{Frontal} & \multicolumn{1}{l|}{7 types} & \multicolumn{1}{l|}{Laser} \\
		\multicolumn{1}{|l|}{3D-TEC~\cite{twinsvijayan2011}} & \multicolumn{1}{c|}{214} & \multicolumn{1}{c|}{428} & \multicolumn{1}{c|}{2} & \multicolumn{1}{c|}{Frontal} & \multicolumn{1}{l|}{None} & \multicolumn{1}{l|}{Laser} \\
		\multicolumn{1}{|l|}{ND-2006~\cite{faltemier2007}} & \multicolumn{1}{c|}{422} & \multicolumn{1}{c|}{9,443} & \multicolumn{1}{c|}{Multiple} & \multicolumn{1}{c|}{$\pm15^\circ$} & \multicolumn{1}{l|}{None} & \multicolumn{1}{l|}{Laser} \\
		\hline
		\multicolumn{1}{|l|}{\textbf{TOTAL}} & \multicolumn{1}{c|}{\textbf{1853}} & \multicolumn{1}{c|}{\textbf{31,860}} & \multicolumn{1}{c|}{\textbf{-}} & \multicolumn{1}{c|}{\textbf{-}} & \multicolumn{1}{c|}{\textbf{-}} & \multicolumn{1}{l|}{\textbf{-}} \\
		\hline
		\multicolumn{7}{l}{All datasets excepts GavabDB come with texture maps (RGB face images).} \\
	\end{tabular}%
	\label{tab:ls3dface}%
	\vspace{-0mm}
\end{table}%

\vspace{0mm}
\section{Large-scale 3D Face Test Dataset}
\vspace{0mm}
FRGCv2~\cite{phillips2005} by far still remains the largest 3D face recognition benchmark dataset with $446$ identities. Table~\ref{tab:introtab} shows that there is a huge disparity between 2D and 3D face dataset sizes. In the absence of any alternate means to collect real 3D faces for testing face recognition systems, we propose a protocol for merging the most challenging public datasets and call this dataset ``\textit{LS3DFace}''.  The proposed technique is akin to crawling the web for augmenting 2D datasets and enables us to create a 3D face dataset of 1,853 identities with 31,860 scans. Our dataset enshrines every possible challenging scenario in face recognition and contains extreme variations in expressions, pose, occlusion, missing data, sensor type and similarities of faces in the form of identical twins. Table~\ref{tab:ls3dface} lists the details of \textit{LS3DFace}. For ease of comparison with our method, we provide the gallery and probe lists for various experiments. Since all these datasets are publicly available, other researchers can reproduce our results using our \textit{FR3DNet} model. Moreover, newly released 3D face datasets can be added to make the protocol more challenging.

Since, ND-2006~\cite{faltemier2007} is a superset of FRGCv2~\cite{phillips2005} dataset, we include the scans common in both datasets only once in the \textit{LS3DFace} to avoid repetitions. Furthermore, BU-4DFE~\cite{bu4yin2008} dataset contains 3D video sequences of six expressions per identity. We only retain five frames equally spaced apart for each expression type.

\vspace{0mm}
\section{Evaluation Protocol}
\vspace{0mm}
We first evaluate the affects of input image size and the convolutional kernel size on face recognition accuracy. We train our network  for $50$ epochs  on 500K 3D faces and report validation accuracy on 100K faces using three image sizes ($96$,$160$ and $224$) and kernel sizes of $3$, $5$, $7$ and $9$. The results are shown in Table~\ref{tab:kernelsize}. The improvement in validation accuracy from image size $96\times96$ to $160\times160$ and from kernel size $5$ to $7$ is significant and hence we select these parameter settings for the remaining experiments. Table~\ref{tab:kernelsize} also validates our claim that bigger kernel sizes are more suitable for 3D data.
\begin{table}[h]
	\centering
	\scriptsize
	\setlength{\tabcolsep}{6.0pt}
	\renewcommand{\arraystretch}{1.25}
	\caption{Affect of image size (at $K=3$) and kernel size $K$ (for $160\times160$ image size)  on validations accuracy. The kernel sizes of only the first two Conv layers are changed.}
	\begin{tabular}{|l|c|c|c|c}
		\cline{1-4}
		\textcolor{blue}{Image size} &  $96\times96$ & $160\times160$ & $224\times224$ &  \\ \cline{1-4}
		{Accuracy($\%$)} & {82.27} & \textbf{86.33}$^*$ & {86.85} &  \\ 		\hline 
		\textcolor{blue}{Kernel size} 	& {$K=3$} & {$K=5$} & {$K=7$} & \multicolumn{1}{c|}{$K=9$} \\ \hline
		{Accuracy($\%$)} & {86.33} & {86.60} & \textbf{88.73}$^*$ & \multicolumn{1}{c|}{88.92} \\ 		\hline
		\multicolumn{5}{l}{\textbf{*} - Significant improvement over smaller kernels ($p<<0.001$)} \\
	\end{tabular}%
	\label{tab:kernelsize}%
	\vspace{0mm}
\end{table}%

We feed forward the 3D images (containing depth, azimuth and elevation normal angles) of \textit{LS3DFace} through \textit{FR3DNet} and use the image representations from FC7 as features. The first available neutral scan of each identity is placed in the gallery while the remaining scans are used as probes. Where a neutral scan is not available, we use the first available scan in the gallery\footnote{The file names of the scans used in gallery will be released.}. Face identification is performed by matching the features of a probe with all identities in the gallery and based on minimum cosine distance, an identity is assigned to the probe. We report the results in the form of Cumulative Matching Curves (CMC). In case of face verification, the probe is matched with each claimed identity in the gallery. The result is a binary accept or reject decision based on some threshold applied to the match score. We report the results as ROC curves for varying thresholds of False Acceptance Rate (FAR). 

Our \textit{FR3DNet}$_{FT}$ is fine-tuned on the gallery set mentioned above which contains a single sample per subject. This is a highly challenging scenario but the most practical one in real world. We learn an \textit{N}-way ($N=1853$) classifier and output the classification decision from the final soft-max layer. We compare our results with the state-of-the-art algorithms on each constituent dataset.


\vspace{+2mm}
\noindent{\textbf{Closed World Face Recognition: }} This is a scenario where all the probes are \textit{enrolled} in the gallery. Such probes are referred to as previously \textit{known}. We report face identification and verification results on \textit{LS3DFace} and its constituent datasets. We also compare our closed world results with four state-of-the-art 2D face recognition CNNs (RGB and 3D) as well as four state-of-the-art conventional methods using the same protocol. Note that wherever we report results on the constituent datasets of \textit{LS3DFace}, the gallery always contains all 1,853 identities and not just the identities of that particular dataset. 

\vspace{+2mm}
\noindent{\textbf{Open World Face Identification: }} A real world scenario in face recognition occurs when the probe set contains \textit{unknown} identities which are not enrolled in the gallery.  Open world or open set 3D face recognition has not been studied in the context of a single sample per person gallery. Scheirer \etal~\cite{scheirer2013} and more recently G\"{u}nther \etal~\cite{gunther2017} discuss this problem in the context of a gallery which contains multiple 2D images of a person and where training a classifier on the gallery faces is involved. Following ~\cite{scheirer2013} we define openness ($\Psi$) with a slight change to account for the single sample per person case used in our experiments:
\begin{equation}
\label{eq:openness}
 \Psi = 1 - \sqrt{\dfrac{2\times N_{\rm TargetID}}{N_{\rm TestID}+N_{\rm TargetID}}} 
\end{equation}
where $N_{\rm TargetID}$ and $N_{\rm TestID}$ denote the number of identities in the gallery and probe sets respectively. $\Psi=0$ ($N_{\rm TargetID} = N_{\rm TestID}$) denotes the conventional closed world face recognition  where as $N_{\rm TargetID}=[1,N_{\rm TestID}-1]$ gives varying levels of {openness} for open world face recognition. The robustness of a face recognition system in open world is tested by varying the \textit{Unknown Person Acceptance Rate} (UPAR) denoted by $\tau$. When $\tau=0$ all probes are classified as an unknown identity whereas when $\tau=1$ every probe is assigned a \textit{known} identity. At each $\tau$ the system outputs the Rank-1 identification rate of all probes (both \textit{known} and \textit{unknown}) identified correctly. We report open world face recognition on \textit{LS3DFace} dataset at different openness values and compare the results with the state-of-the-art algorithms.

\begin{table*}[htbp]
	\centering
	\scriptsize
	\setlength{\tabcolsep}{2.0pt}
	\renewcommand{\arraystretch}{1.25}
	\caption{Comparison of face recognition accuracy ($\%$) on \textit{LS3DFace} with state-of-the-art deep and conventional methods. We have used the complete gallery of \textit{LS3DFace} in all experiments reported here.}
	\begin{tabular}{|c|l|c|c|c|c|c|c|c|c|c|c|c|c|}
		\hline
		\multirow{3}[4]{*}{\textbf{Method}} & \multirow{2}[4]{*}{\textbf{Model \textbackslash}} & \multirow{3}[4]{*}{\textbf{Modality}} & \multicolumn{11}{c|}{\textbf{Gallery of LS3DFace}} \\
		\cline{4-14}          &     &       & \textbf{LS3DFace} & \textbf{FRGC} & \textbf{BU3DFE} & \textbf{BU4DFE} & \textbf{Bosphorus} & \textbf{CASIA} & \textbf{GavabDB} & \textbf{TexasFRD} & \textbf{3D-TEC} & \textbf{UMBDB} & ND-2006 \\
		&   \textbf{Technique}     &        & \textbf{This paper} & \multicolumn{1}{l|}{\cite{phillips2005}} & \multicolumn{1}{c|}{\cite{yin2006} } & \multicolumn{1}{c|}{\cite{bu4yin2008}} & \multicolumn{1}{c|}{ \cite{savran2008} } & \multicolumn{1}{c|}{\cite{casia} } & \multicolumn{1}{c|}{\cite{gavabmoreno2004} } & \multicolumn{1}{c|}{\cite{texasgupta2010} } & \multicolumn{1}{c|}{\cite{twinsvijayan2011} } & \multicolumn{1}{c|}{\cite{umbcolombo2011} } & \cite{faltemier2007} \\
		\hline
		\multirow{6}[4]{*}{\begin{sideways}CNN\end{sideways}} & {GoogleNet~\cite{googleszegedy2015}} & RGB   & 53.97 & 21.51 & 50.76 & 65.41 & 63.44 & 85.91 & - & 53.08 & 79.95 & 65.78 & 24.14 \\
		& {Resnet152~\cite{resnethe2016}} & RGB   & 15.05 & 13.53 & 8.04  & 9.64  & 7.05  & 52.85 & - & 20.94 & 72.66 & 34.08 & 10.92 \\
		& {VGG-Face~\cite{parkhi2015}} & RGB   & 90.85 & 87.92 & 97.68 & 96.51 & 96.39 & 94.18 &- & 99.73 & 83.30 & 81.54 & 82.86 \\
		\cline{2-14}          & {GoogleNet~\cite{googleszegedy2015}} & 3D    & 38.66 & 35.54 & 46.56 & 41.88 & 26.81 & 50.81 & 66.56 & 67.59 & 67.29 & 47.66 & 30.81 \\
		& {Resnet152~\cite{resnethe2016}} & 3D   & 12.49 & 14.40 & 5.80  & 10.13 & 3.84  & 25.34 & 44.26 & 16.25 & 60.98 & 22.20 & 12.08 \\
		& {VGG-Face~\cite{parkhi2015}} & 3D   & 61.20 & 62.42 & 71.16 & 53.17 & 48.14 & 71.95 & 77.38 & 85.58 & 78.04 & 67.48 & 60.81 \\
		\hline
		\multirow{4}[1]{*}{\begin{sideways}Conventional\end{sideways}} & \multicolumn{1}{l|}{MMH~\cite{mian2007}} & 3D + 2D & 83.08 & 89.37 & 88.50 & 84.93 & 85.10 & 85.24 & 86.64 & 85.67 & 80.85 & 77.32 & 86.71 \\
		& 3D Keypoint~\cite{mian2008} & 3D    & 81.76 & 86.59 & 85.14 & 82.50 & 82.64 & 81.38 & 84.41 & 84.99 & 75.63 & 71.68 & 82.30 \\
		& R3DM~\cite{gilani2017a} & 3D    & 82.89 & 87.50 & 87.13 & 83.21 & 86.06 & 84.51 & 85.60 & 85.47 & 78.27 & 77.11 & 84.84 \\
		& K3DM~\cite{gilani2017b} & 3D    & 84.67 & 89.50 & 89.24 & 86.05 & 88.60 & 85.35 & 87.90 & 86.13 & 79.55 & 78.64 & 87.77 \\ \hline \hline
		\multirow{2}[1]{*}{\begin{sideways}CNN\end{sideways}} & \textbf{FR3DNet} & 3D &   95.51 & 97.06 & 98.64 & 95.53 & 96.18 & 98.37 & 96.39 & 100.00 & 97.90 & 91.17 & 95.62 \\
		& \textbf{FR3DNet}$_{FT}$ & \textbf{3D} & \textbf{98.75} & \textbf{99.88} & \textbf{99.96} & \textbf{98.04} & \textbf{100.00} & \textbf{99.74} & \textbf{99.70} & \textbf{100.00} & \textbf{99.12} & \textbf{97.20} & \textbf{99.13} \\ \hline
	\end{tabular}%
	\label{tab:dfn_ls3db}%
	\vspace{-0mm}
\end{table*}%

\section{Results and Analysis}

\subsection{Closed World Face Recognition}

\begin{figure*}[h]
	\begin{minipage}[b]{0.31\linewidth}
		\centering
		\includegraphics[trim = 5pt 0pt 25pt 10pt, clip, width=1\linewidth]{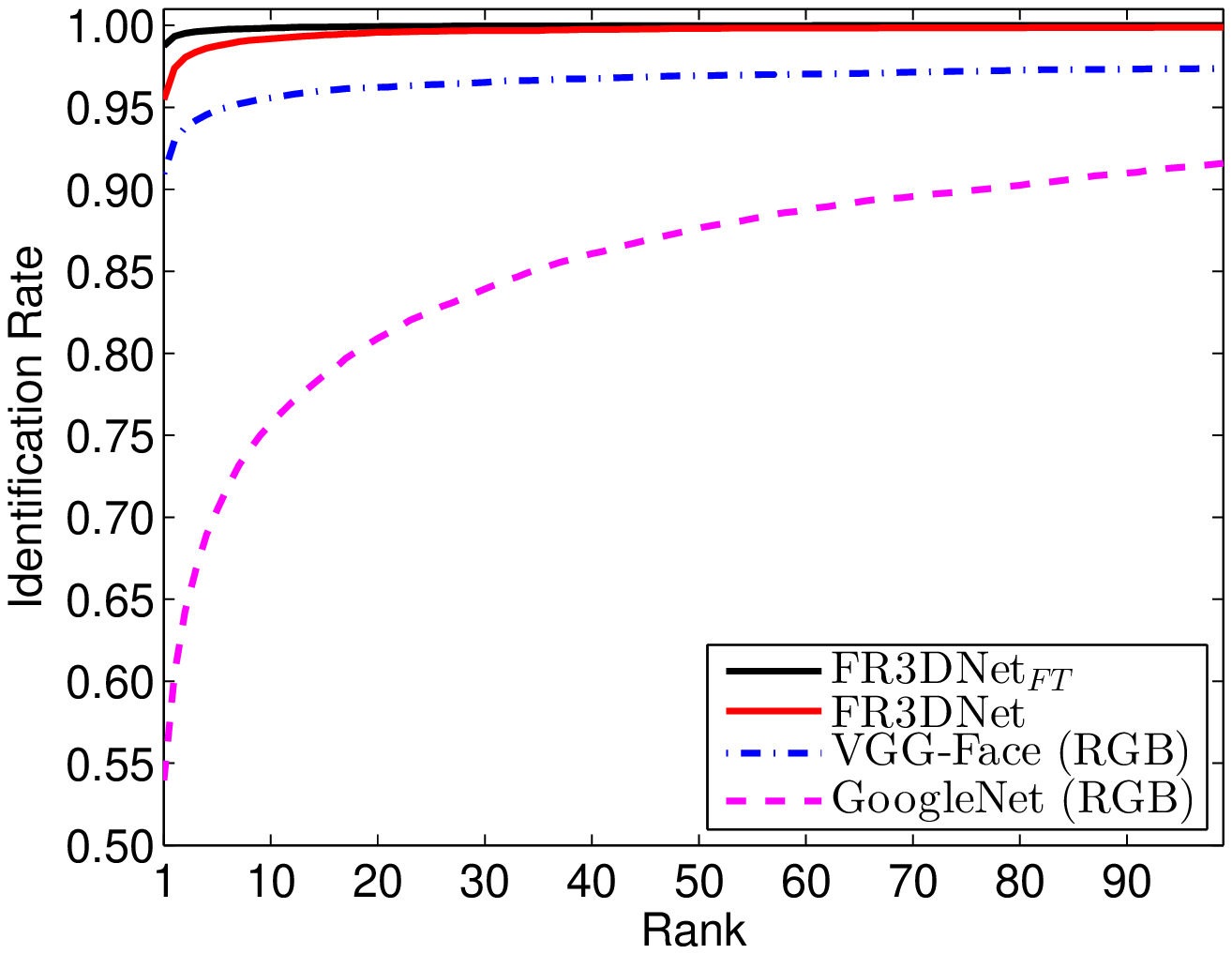}
		\centerline{\small{(a) }}\medskip
	\end{minipage}
	\hfill
	\begin{minipage}[b]{0.31\linewidth}
		\centering
		\includegraphics[trim = 5pt 0pt 25pt 10pt, clip, width=1\linewidth]{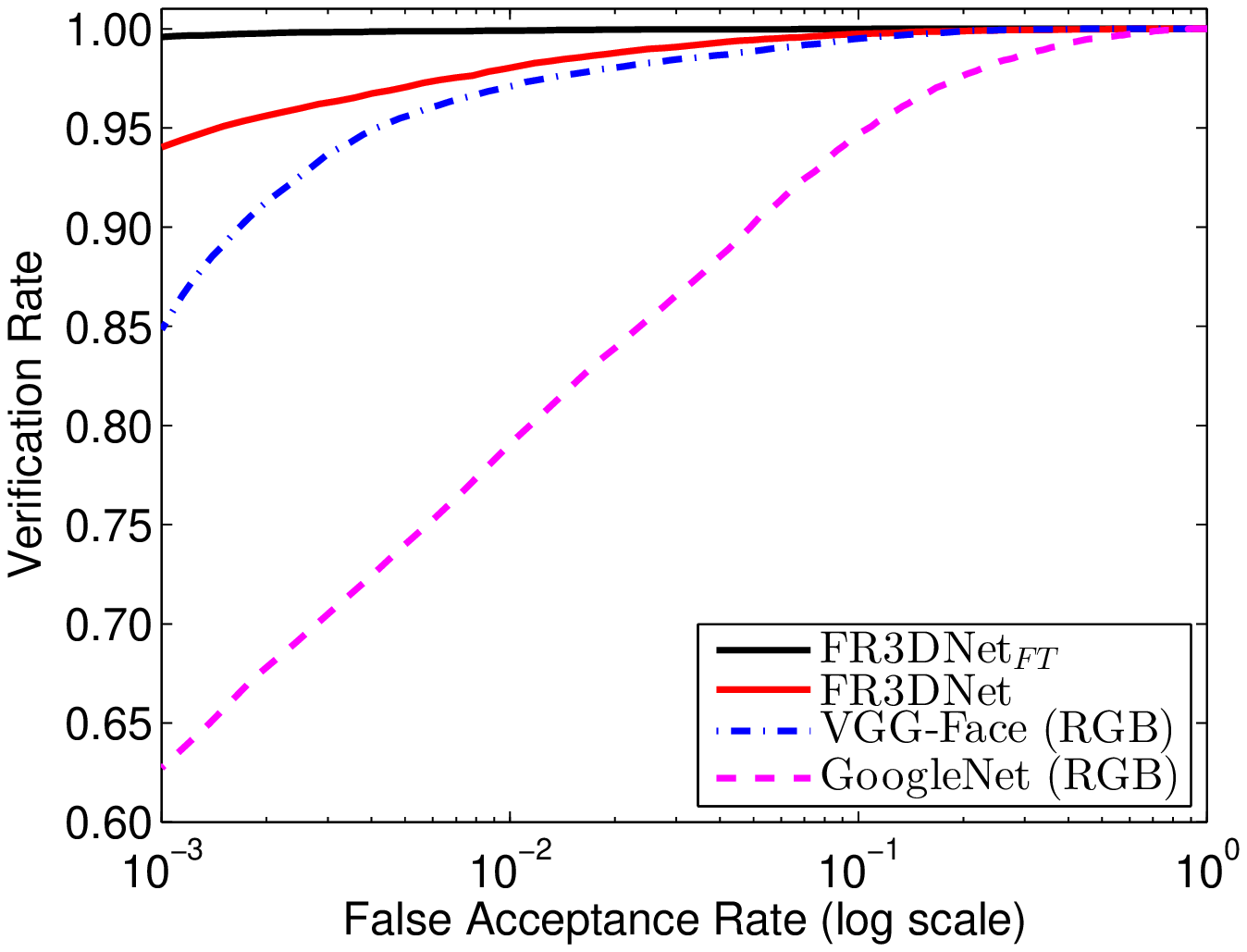}
		\centerline{\small{(b)}}\medskip
	\end{minipage}
	\hfill
	\begin{minipage}[b]{0.31\linewidth}
		\centering
		\includegraphics[trim = 5pt 0pt 25pt 10pt, clip, width=1\linewidth]{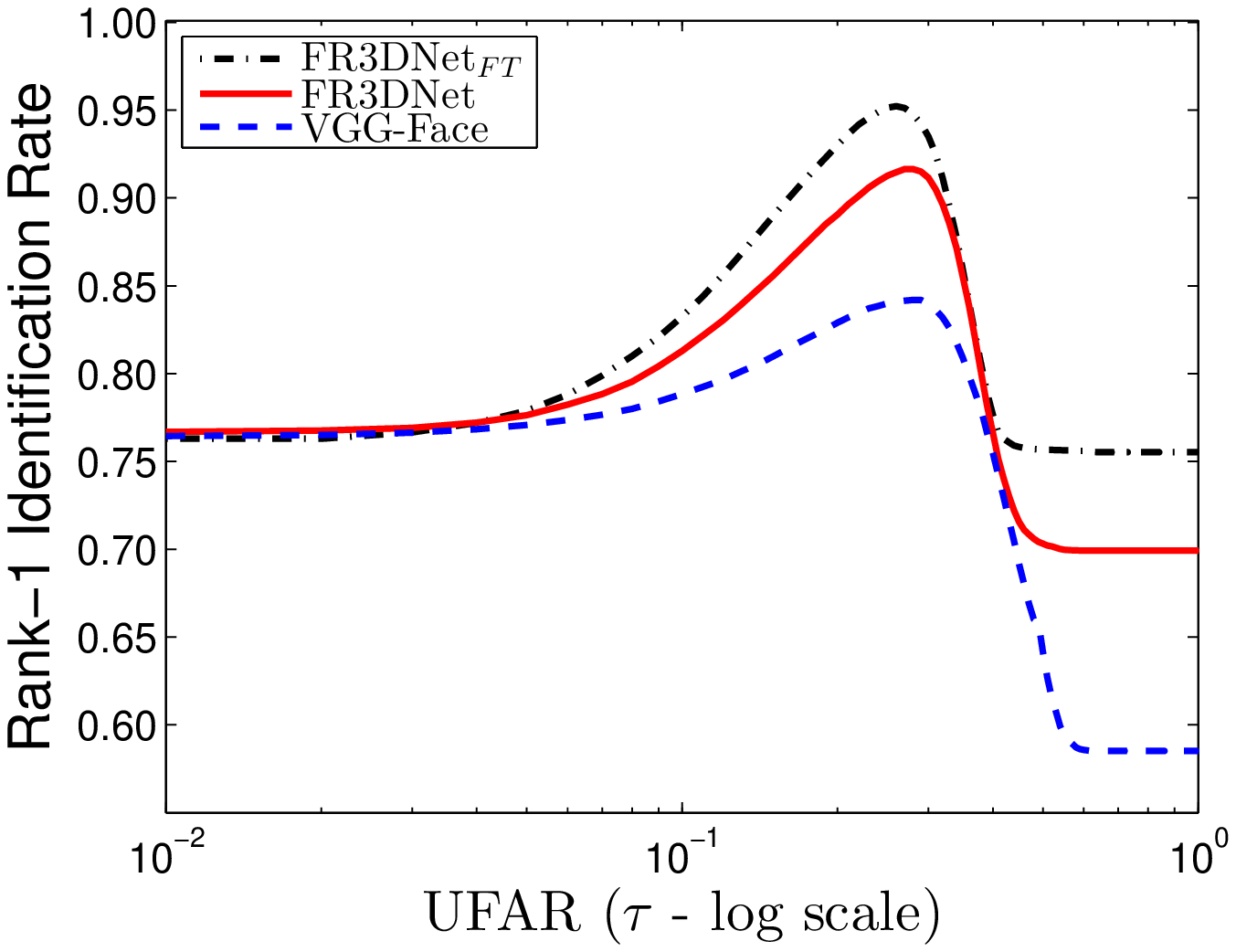}
		\centerline{\small{(c)}}\medskip
	\end{minipage}
	\vspace{0mm}
	\caption{\small{Comparison of closed world (a) CMC and (b) ROC curves with state-of-the-art algorithms on \textit{LS3DFace} dataset. (c) Comparison of open world Rank-1 identification rates at varying thresholds of unknown face acceptance rate (UFAR). The gallery of \textit{LS3DFace} has been reduced by $50\%$. Note that \textit{FR3DNet} performs better than VGG-Face by a margin of $10\%$. Curves for only the best performing networks from Table~\ref{tab:dfn_ls3db} are shown. }}
	
	\label{fig:cmcroc_ls3db}
	\vspace{0mm}
\end{figure*}

Table~\ref{tab:dfn_ls3db} details the closed world Rank-1 identification results on \textit{LS3DFace} and compares them with the state-of-the-art deep and conventional methods. We perform 2D face recognition on the RGB images that accompany the datasets. The main conclusions that can be drawn from these results is that 3D face has more to offer in terms of correctly identifying a person and 3D face recognition results are superior than its 2D counterpart. Note that the conventional methods that report near saturated results on small 3D face datasets, fail to achieve high accuracies on the large \textit{LS3DFace} dataset. This shows that increasing the gallery size has a strong inverse affect on the performance of these algorithms. \textit{FR3DNet} outperforms state-of-the-art conventional 3D face recognition algorithms by more than $14$\% and the best 2D face recognition method by $4.7$\%. In a single sample per face gallery scenario, only 3D faces have the advantage to generate more training data to fine-tune any network. \textit{FR3DNet$_{FT}$} outperforms VGG-Face by 8\% and the margins are more significant (over 15\%) when the probes are more challenging as in the cases of 3D-TEC (twins) and UMBDB datasets.
We compare the our CMCs and ROCs with VGG-Face and GoogleNet in Figure~\ref{fig:cmcroc_ls3db}.

\begin{table}[tbp]
	\centering
	\scriptsize
	\setlength{\tabcolsep}{2.0pt}
	\renewcommand{\arraystretch}{1.25}
	\caption{Comparison of Rank-1 recognition accuracy ($\%$) with the state-of-the-art methods on constituent datasets of \textit{LS3DFace}. Note that FR3DNet$_{FT}$ uses the full gallery of \textit{LS3DFace} dataset. }
	\begin{tabular}{|l|c|c|c|c|c|c|c|c|c|c|}
		\hline
		\multicolumn{1}{|c|}{\textbf{Method/Dataset}} &  
		{\begin{sideways}\textbf{FRGCv2} \cite{phillips2005}\end{sideways}} &  
		{\begin{sideways}\textbf{BU3DFE} \cite{yin2006} \end{sideways}} &  
		{\begin{sideways}\textbf{BU4DFE} \cite{bu4yin2008}\end{sideways}} &  
		{\begin{sideways}\textbf{Bosphorus} \cite{savran2008} \end{sideways}} &  
		{\begin{sideways}\textbf{CASIA} \cite{casia} \end{sideways}} &  
		{\begin{sideways}\textbf{GavabDB} \cite{gavabmoreno2004} \end{sideways}} &  
		{\begin{sideways}\textbf{TexasFRD} \cite{texasgupta2010} \end{sideways}} &  
		{\begin{sideways}\textbf{3D-TEC} \cite{twinsvijayan2011} \end{sideways}} &  
		{\begin{sideways}\textbf{UMBDB} \cite{umbcolombo2011} \end{sideways}} &  
		{\begin{sideways}\textbf{ND-2006} \cite{faltemier2007}\end{sideways}} \\ \hline
		Xu \etal\cite{casia} & -     & -     & -     & -     & 83.9  & -     & -     & -     & -     & - \\
		Kim \etal\cite{kim2017} & -     & 95.0  & -     & 99.2  & -     & -     & -     & -     & -     & - \\
		Faltemier \etal~\cite{faltemier2007} & -     & -     & -     & -     & -     & -     & -     & -     & -     & 82.8 \\
		Gupta \etal~\cite{texasgupta2010} & -     & -     & -     & -     & -     & -     & 97.9  & -     & -     & - \\
		Al-Osaimi \etal~\cite{al2016} & 97.8  & -     & -     & -     & -     & -     & -     & 97.2  & -     & - \\
		Li \etal~\cite{li2014,li2014b} & 96.3  & 92.2  & -     & 96.6  & -     & -     & -     & 96.7  & -     & - \\
		Lei \etal~\cite{lei2016} & 96.3  & 94.0  & -     & -     & -     & 96.3  & -     & -     & 73.1  & - \\
		Mian \etal~\cite{mian2007} & 96.2  & 95.9  & 94.2  & 96.4  & 82.5  & 95.4  & 98.0  & 95.9  & 69.3  & 95.3 \\
		Gilani \etal~\cite{gilani2017b} & 98.5  & 96.2  & 96.0  & 98.6  & 85.4  & 96.5  & 98.1  & 92.6  & 78.6  & 96.8 \\
		\hline \hline
		
		\boldmath{}\textbf{ FR3DNet$_{FT}$}\unboldmath{} & \textbf{99.9} & \textbf{99.9} & \textbf{98.0} & \textbf{100.0} & \textbf{99.7} & \textbf{99.7} & \textbf{100.0} & \textbf{99.1} & \textbf{97.2} & \textbf{99.1} \\
		\hline
	\end{tabular}%
	\label{tab:cw_comp_sota}%
\end{table}%

A straightforward comparison of \textit{FR3DNet}$_{FT} $ on the individual datasets in Table~\ref{tab:cw_comp_sota} shows that our deep network fine-tuned on a single scan per person outperforms  the state-of-the-art conventional algorithms. The results for all other methods are reported from their original papers. Note that these results are biased against \textit{FR3DNet}$_{FT}$ which tests each probe against the full gallery of \textit{LS3DFace} dataset whereas other methods test the probes against only the gallery identities of that particular dataset. Results are remarkable especially on the UMBDB~\cite{umbcolombo2011} (containing occlusions) and CASIA datasets where our network outperforms the nearest competitors by 18.6\% and 14.3\% respectively. 


\begin{table}[tb]
	\vspace{-4mm}
	\centering
	\scriptsize
	\setlength{\tabcolsep}{4.0pt}
	\renewcommand{\arraystretch}{1.25}
	\caption{Comparison of average open world Rank-1 recognition accuracy($\%$) with the state-of-the-art 2D face recognition networks at varying levels of openness (see Equation~\ref{eq:openness}). The standard deviation of the accuracies over ten random folds is less than $1\%$.}
	\vspace{2mm}
	\begin{tabular}{|l|c|c|c|c|c|c|}
		\hline
		\textbf{Openness} & \textbf{4\%} & \textbf{9\%} & \textbf{15\%} & \textbf{23\%} & \textbf{33\%} & \textbf{50\%} \\
		\textbf{num Unkown IDs} & 272   & 542   & 810   & 1072  & 1321  & 1592 \\     \hline
		
		GoogleNet~\cite{googleszegedy2015} & 43.70 & 38.54 & 34.64 & 28.41 & 23.70 & 19.28 \\
		VGG-Face~\cite{parkhi2015} & 86.61 & 80.56 & 71.31 & 63.52 & 54.54 & 48.44 \\    \hline     \hline
		\textbf{FR3DNet} & 92.41 & 88.63 & 78.42 & 71.51 & 64.10 & 57.02  \\
		\textbf{ FR3DNet$_{FT}$} & 97.22 & 91.94 & 83.72 & 77.80 & 70.21 & 61.20  \\
		\hline
	\end{tabular}%
	\label{tab:owR1}%
\end{table}%

\subsection{Open World Face Identification}

Unlike other methods~\cite{scheirer2013,gunther2017}, we do not use classifiers to train on a subset of gallery faces since we perform face recognition with a single sample per person in the gallery. Figure~\ref{fig:cmcroc_ls3db}(c) shows Rank-1 identification results when half  of the gallery ($925$) identities are removed to simulate an open world scenario. Hence, the probes belonging to these identities are \textit{unknown}. The curves demonstrate the detection power of \textit{FR3DNet} which outperforms VGG-Face (RGB) by a significant margin. In Table~\ref{tab:owR1} we report the average Rank-1 recognition rate over $\tau=[0,1]$ for varying levels of openness(See Equation~\ref{eq:openness}). For both experiments we performed  ten random fold selection of \textit{unknowns} and the figures presented in 
are the mean results of ten random folds. The standard deviations in all cases was less than $1\%$.

\section{Conclusion}
This paper bridges the vast gap between research advancements in 2D and 3D face recognition algorithms especially in the context of deep learning. It proposes a technique to generate millions of 3D facial images of unique identities by simultaneously interpolating between the facial identity and facial expression spaces. Additional factors such as subtle variations in facial shape, major variations in facial shape, camera viewpoint and self occlusions are introduced to generate a training dataset of 3.1M scans of 100K identities. A purpose designed 3D face recognition CNN is proposed and trained from scratch on this dataset. To test the network, existing 3D face datasets are merged and comparative results are reported on the largest 3D face dataset to date. The proposed training and test datasets are several orders of magnitude larger than the existing 3D datasets reported in the literature. The proposed \textit{FR3DNet} outperforms the state-of-the-art 3D as well as 2D face recognition algorithms in closed and open world recognition scenarios.

\section*{Acknowledgments}
	This research was supported by ARC grant DP160101458. The Titan Xp used for this research was donated by NVIDIA Corporation.

\end{document}